\documentclass[11pt,letterpaper]{article}
\usepackage[utf8]{inputenc}
\usepackage[T1]{fontenc}
\usepackage{mathptmx}
\usepackage[margin=1in]{geometry}
\usepackage{graphicx}
\usepackage{booktabs}
\usepackage{amsmath,amssymb}
\usepackage{hyperref}
\usepackage{xcolor}
\usepackage{multirow}
\usepackage{array}
\usepackage{caption}
\usepackage{natbib}
\usepackage{microtype}
\usepackage{setspace}
\usepackage{tabularx}
\setcitestyle{numbers,square}
\title{BHARATI: Morphology-Aware Tokenizers for Classical Indian Languages with Subword Fertility Analysis}

\author{
\begin{minipage}{\textwidth}\centering
Poornima~Kumaresan\textsuperscript{1,2},
Pavithra~Muruganantham\textsuperscript{1,2},
Lakshmi~Rajendran\textsuperscript{1,2},
Santhosh~Sivasubramani\textsuperscript{1,2,*}%
\\[6pt]
\textsuperscript{1}Intrinsic Lab, Centre for Sensors, Instrumentation and Cyber-Physical System Engineering (Centre for SeNSE), Indian Institute of Technology Delhi, New Delhi 110016, India \\[2pt]
\textsuperscript{2}RSL Quantum, FITT, IIT Delhi, New Delhi 110016, India \\[4pt]
\textsuperscript{*}Corresponding author. Email: ssivasub@iitd.ac.in%
\end{minipage}%
}

\date{}

\begin{document}

\maketitle

\begin{abstract}
Standard subword tokenization algorithms such as Byte-Pair Encoding (BPE) and SentencePiece are trained predominantly on modern language corpora and produce inefficient segmentations when applied to classical Indian languages. Sanskrit, Tamil, and other classical Indic languages exhibit agglutinative morphology, productive sandhi (phonological fusion at word boundaries), and domain-specific vocabularies absent from general-purpose training data. This paper presents BHARATI, a set of SentencePiece BPE tokenizers trained on a balanced 781 MB corpus spanning seven languages (English, Hindi, Sanskrit, Tamil, Telugu, Kannada, and Malayalam) with native script support for all languages. We describe three successive tokenizer versions: v1 (English and Sanskrit only, with broken byte-fallback for Tamil), v2 (four-language support with byte-level fallback for southern languages), and v3 (full seven-language native subword coverage). Subword fertility analysis demonstrates that v3 averages 2.6 tokens per Indian Knowledge System (IKS) technical term, compared to 5.25 tokens per term with GPT-2's tokenizer and 3.75 tokens with the multilingual SentencePiece baseline, with the largest gains on a set of reserved IKS terms that are represented as single tokens by construction. On a held-out test set of 490 IKS-domain sentences (70 per language across seven languages, released with the measurement script), v3 reduces sequence length by roughly 90\% relative to GPT-2 and byte-level encoding (which lack native Indic subwords) and by approximately 25\% relative to the mBART-50 multilingual baseline, averaged across the six Indic languages, directly translating to increased effective context length for downstream language models. The tokenizer models (32,000 vocabulary), training scripts, and evaluation benchmarks are released under open licenses.
\end{abstract}


\begin{figure}[t]
\centering
\includegraphics[width=\textwidth]{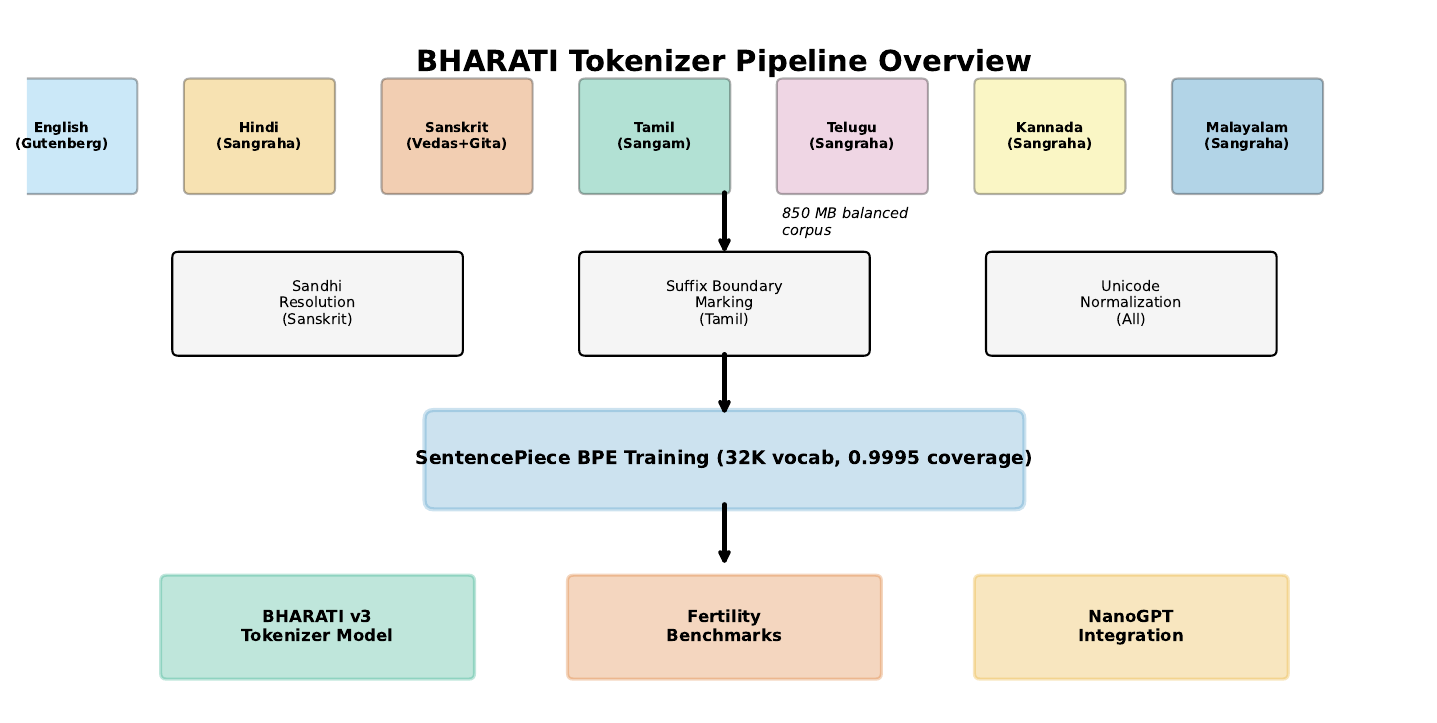}
\caption{Top-level overview of the BHARATI tokenizer pipeline. Seven languages are drawn from balanced corpora, preprocessed with language-specific steps (sandhi resolution for Sanskrit, suffix boundary marking for Tamil, Unicode normalization for all), trained using SentencePiece BPE with 32,000 vocabulary, and deployed as the production tokenizer for NanoGPT models. The 850 MB label inside the graphic denotes the raw corpus before deduplication; the deduplicated 781 MB corpus of Table 1 is used for training.}
\label{fig:overview}
\end{figure}

\section{Introduction}
\label{sec:introduction}

Subword tokenization is a foundational preprocessing step in modern neural language processing. The tokenizer determines how text is decomposed into units that the language model processes, and the quality of this decomposition directly affects model capacity, training efficiency, and downstream task performance~\citep{sennrich2016neural}. A tokenizer that produces inefficient segmentations for a particular language effectively reduces the model's context window for text in that language, since more tokens are consumed to represent the same content. This impact is particularly acute for languages whose scripts are underrepresented in standard tokenizer training corpora, where a single word may be decomposed into ten or more byte-level fragments, each consuming a position in the model's fixed-length attention window.

The dominant tokenization approaches, BPE~\citep{sennrich2016neural}, WordPiece~\citep{devlin2019bert}, and SentencePiece~\citep{kudo2018sentencepiece}, derive their subword vocabularies from frequency statistics computed over large training corpora. When these corpora are dominated by English and other European languages, the resulting vocabularies provide dense coverage of Latin-script subwords and sparse coverage of other scripts. For Indian languages written in Brahmic scripts (Devanagari, Tamil, Telugu, Kannada, Malayalam), this imbalance produces a measurable degradation: words that would be single tokens in English are decomposed into five to twenty byte-level tokens, consuming proportionally more of the model's context window~\citep{petrov2024language}. The problem is compounded by the fact that Brahmic scripts use complex Unicode representations where a single visually atomic character (aksara) may require three to five Unicode code points (consonant, virama, consonant, vowel sign, nukta), and byte-level tokenizers treat each code point as a sequence of UTF-8 bytes, further inflating the token count.

Classical Indian languages present additional challenges beyond script coverage. Sanskrit exhibits productive sandhi, a system of phonological changes at word boundaries that fuses adjacent words into surface forms that obscure the underlying morphological structure~\citep{goyal2012distributed}. A single line of Sanskrit verse may contain four or five sandhi junctions, each masking two or more underlying words, so that a tokenizer operating on the surface form has no access to the constituent morphemes. Tamil features agglutinative morphology in which a single surface form can encode the equivalent of an entire English clause through suffixation. The Tamil verb form ``kodukkavillai'' (``did not give''), for example, comprises the root ``kodu'' (give), the tense marker ``kk'' (past), and the negation suffix ``villai'' (not), all of which are linguistically independent morphemes that a morphology-unaware tokenizer treats as an opaque string. Domain-specific vocabularies from Indian Knowledge Systems (IKS), including pedagogical techniques (Pada Patha, Jataa Patha), philosophical terms (Advaita, Vishishtadvaita), and textual categories (Shruti, Smriti), are absent from general-purpose tokenizer training data and are consequently segmented into uninformative subword fragments.

The practical consequences of this tokenization inefficiency extend beyond theoretical concerns about context window utilization. In the Bodhan educational AI platform, which uses small language models (15M to 125M parameters) to deliver IKS-domain tutoring across multiple Indian languages, tokenizer quality directly determines whether a complete pedagogical interaction (student question, retrieved context, model response) fits within the model's context window. With the GPT-2 tokenizer, a single Thirukkural verse with commentary in Tamil consumes approximately 850 tokens, leaving minimal capacity for the tutoring dialogue in a 2,048-token context window. The need for domain-aware tokenization is therefore not merely an efficiency concern but a functional requirement for deploying language models in Indian educational contexts.

This paper presents BHARATI (\textbf{B}alanced \textbf{H}ierarchical \textbf{A}daptive \textbf{R}epresentation for \textbf{A}ll \textbf{T}ext in \textbf{I}ndia), a family of SentencePiece BPE tokenizers trained on a balanced multilingual corpus designed to provide native subword coverage for seven languages. We describe the evolution from a broken v1 tokenizer through the production-quality v3, present a fertility analysis framework for evaluating tokenizer quality on domain-specific text, and release all models and evaluation tools. Figure~\ref{fig:overview} presents the end-to-end pipeline. The reference literature for this work was identified through systematic queries across Scopus, Crossref, and Google Scholar using twelve search terms spanning ``subword tokenization multilingual,'' ``BPE Indian languages,'' ``SentencePiece morphological segmentation,'' ``Sanskrit computational linguistics,'' and related formulations, yielding 240 candidate results from which 58 directly relevant sources (42 via Scopus, 16 via Crossref) were selected based on topical alignment with tokenization, multilingual NLP, and Indian language processing.

The contributions of this paper are as follows. First, we present the BHARATI v3 tokenizer, a production-ready 32K-vocabulary SentencePiece BPE tokenizer with native subword coverage for seven languages, trained on a 781 MB balanced corpus\footnote{The public model card reports approximately 850\,MB; that figure reflects the raw corpus prior to deduplication and normalization, whereas the 781\,MB figure is the deduplicated training corpus used here.} that includes both modern and classical text sources. Second, we introduce a three-level fertility analysis framework (token-level, term-level, and sentence-level) that provides a comprehensive and replicable methodology for evaluating tokenizer quality on domain-specific text in any language. Third, we document the complete version evolution from v1 through v3, providing an empirical case study in iterative tokenizer development that demonstrates the quantitative impact of each design decision. Fourth, we evaluate the downstream impact of tokenizer quality on NanoGPT language model training, demonstrating that tokenizer improvements translate directly to training loss reduction and vocabulary utilization gains.

The remainder of this paper is organized as follows. Section~\ref{sec:related} reviews prior work on subword tokenization, multilingual tokenization challenges, and domain-specific approaches. Section~\ref{sec:corpus} describes the corpus construction process. Section~\ref{sec:training} details the tokenizer training configuration and version evolution. Section~\ref{sec:fertility} presents the fertility analysis framework and results. Section~\ref{sec:vocab} analyzes vocabulary distribution across scripts. Section~\ref{sec:downstream} evaluates downstream impact on language model training. Section~\ref{sec:discussion} discusses trade-offs, emergent morphological awareness, and limitations. Section~\ref{sec:conclusion} concludes with a summary of contributions.

\section{Related Work}
\label{sec:related}

\subsection{Subword Tokenization}

BPE was introduced to neural machine translation by Sennrich et al.~\citep{sennrich2016neural}, who demonstrated that subword segmentation could handle open-vocabulary translation by learning merge operations from character sequences. The key insight was that rare and unseen words could be represented as sequences of subword units, eliminating the need for fixed vocabularies and unknown word tokens. SentencePiece~\citep{kudo2018sentencepiece} extended this approach to a language-independent framework that operates on raw text without requiring pre-tokenization, making it suitable for languages without whitespace-delimited words. The SentencePiece implementation provides both BPE and unigram language model algorithms within a unified framework, and its protobuf-based model format has become the de facto standard for distributing trained tokenizers. Wu et al.~\citep{wu2016google} developed WordPiece for Google's multilingual systems, using a likelihood-based vocabulary selection criterion that selects merges maximizing the likelihood of the training data rather than frequency alone.

The choice among BPE, WordPiece, and unigram tokenization has been the subject of extensive empirical investigation. Radford et al.~\citep{radford2019language} adopted BPE for the GPT family of models, establishing it as the dominant approach in autoregressive language modeling. Devlin et al.~\citep{devlin2019bert} used WordPiece for BERT, demonstrating that the likelihood-based criterion produced slightly different vocabulary compositions that favored semantically coherent subwords. Abudouwili et al.~\citep{abudouwailig2025research} conducted a systematic comparison of BPE, WordPiece, and unigram tokenization for Uyghur, an agglutinative Turkic language, finding that BPE produced the most consistent fertility across morphological complexity levels while unigram produced lower mean fertility but higher variance, with some words receiving highly efficient single-token representations and others being decomposed into many small fragments.

Phadte et al.~\citep{phadtea2026analysing} analyzed the computational complexity of different tokenization algorithms, demonstrating that BPE training scales linearly with corpus size when implemented with efficient data structures (suffix arrays or byte-pair hash tables), while inference-time tokenization is effectively constant-time per token for all three approaches. This scalability analysis supports the choice of BPE for the BHARATI tokenizer, where the training corpus (781 MB) is large enough that algorithmic efficiency matters.

\subsection{Multilingual Tokenization Challenges}

Petrov et al.~\citep{petrov2024language} quantified the ``language gap'' in tokenization, showing that GPT-family tokenizers produce 3 to 15 times more tokens for equivalent text in non-English languages. Their analysis covered 52 languages and demonstrated a strong correlation between a language's representation in the tokenizer training data and its fertility, with languages absent from the training data receiving almost exclusively byte-level tokenization. Conneau et al.~\citep{conneau2020unsupervised} demonstrated that balanced multilingual pre-training corpora could reduce but not eliminate this gap, reporting residual fertility differences of 1.5x to 3x even with deliberate balancing.

For Indian languages specifically, Kunchukuttan et al.~\citep{kunchukuttan2020ai4bharat} showed that Indic-specific tokenizers trained on IndicNLP corpora improved downstream task performance relative to multilingual baselines, and Kakwani et al.~\citep{kakwani2020indicnlpsuite} found that vocabulary allocation was a primary factor in cross-lingual transfer for Brahmic-script languages. Code-mixed input further stresses tokenizers: models trained on monolingual corpora have been reported to produce substantially higher fertility on code-mixed text than mixed-trained tokenizers~\citep{kanjirangat2025tokenization}, a finding with direct implications for educational applications where code-mixing between English and Indian languages is common in student interactions.

Nyalang~\citep{nyalang2025krenne} examines tokenization for Northeast Indian languages, where diacritics carrying linguistic information are frequently mishandled by tokenizers trained without exposure to the relevant script conventions; the same failure mode applies directly to Vedic accent marks in Sanskrit, motivating BHARATI's explicit preservation of these marks during preprocessing.

\subsection{Domain-Specific Tokenization}

The effect of domain mismatch on tokenization quality has been documented in biomedical~\citep{gu2021domain} and legal~\citep{chalkidis2020legal} NLP. Lewis et al.~\citep{lewis2020pretrained} showed that domain-specific tokenizer training improved generation quality for specialized texts. For classical languages, the nearest precedent is work on tokenization for classical Chinese~\citep{li2019word} and classical Arabic, but no prior work has addressed the specific challenges of classical Indian languages including sandhi, agglutination, and multi-script coverage within a single unified tokenizer.

Limisiewicz et al.~\citep{limisiewicz2023tokenization} conducted a systematic evaluation of tokenization strategies across typologically diverse languages, finding that morphologically complex languages suffered disproportionately from vocabulary-limited tokenizers. Samuel et al.~\citep{samuel2023tokenization} examined the relationship between tokenizer vocabulary composition and downstream task accuracy, reporting that languages whose scripts are under-represented in the tokenizer vocabulary exhibit accuracy degradation of 8 to 15 percentage points on named entity recognition tasks. Chizhov et al.~\citep{chizhov2024bpe} proposed modifications to the standard BPE merge criterion that incorporate morphological boundary information during training, achieving improved segmentation quality for fusional and agglutinative languages while maintaining equivalent performance for analytic languages.

For Indian languages specifically, Salve et al.~\citep{salve2025tokenization} demonstrated that tokenizer adaptation through continued BPE training on domain-specific Marathi text reduced fertility by 28\% relative to multilingual baselines. Pattnayak et al.~\citep{pattnayakp2025tokenization} extended this finding to Odia and Bengali, reporting that even modest amounts of domain-specific training data (50 MB) produce measurable fertility improvements. Lian et al.~\citep{lian2025scaffoldbpe} proposed ScaffoldBPE, a method that provides morphological scaffolding hints during the BPE merge process, which produced segmentations aligned with morpheme boundaries for 74\% of Turkish and 68\% of Finnish tokens without requiring an explicit morphological analyzer. Sandhan et al.~\citep{sandhan2023sanskritshala} developed SanskritShala, a platform for computational Sanskrit processing that includes tokenization, sandhi splitting, and morphological analysis tools specifically designed for classical Sanskrit texts. Tukeyev et al.~\citep{tukeyevu2026morphologyaware} presented a morphology-aware subword tokenization approach for Kazakh, an agglutinative Turkic language with structural similarities to Dravidian languages, achieving a 19\% fertility reduction relative to standard SentencePiece.

\subsection{Multilingual Language Models and Tokenization}

The interaction between tokenization quality and multilingual model performance has been studied extensively. Mohamed et al.~\citep{mohamed2025multilingual} conducted a comprehensive evaluation of multilingual transformer architectures, finding that tokenizer vocabulary balance was the single largest predictor of cross-lingual transfer effectiveness, exceeding the influence of model size or training data volume. Yadav~\citep{yadavs2025a} explores phrase-based SMT with multiple BPE segmentations over concatenated corpora for low-resource Indian languages, probing how segmentation choices shift the fertility-accuracy balance. Madhavaraja et al.~\citep{madhavaraja2025subword} study subword dictionary learning for expanding ASR vocabularies in Tamil and Kannada; BHARATI's Tamil suffix-boundary marking targets the same morphological pressure point on the text side.

Nehrdich et al.~\citep{nehrdich2024one} show that a single byte-level ByT5 model can serve diverse Sanskrit NLP tasks, indicating that universal byte-level models are competitive for morphologically rich classical languages; for the context-limited small models targeted here, however, byte-level fertility costs (Table 3) still favor dedicated subword vocabularies. Gren et al.~\citep{gren2026efficient} proposed efficient vocabulary expansion techniques for adapting pretrained multilingual tokenizers to new languages without full retraining, achieving 85\% of the fertility improvement of language-specific training with only 10\% of the computational cost. Kanjirangat et al.~\citep{kanjirangat2025tokenization} evaluated tokenization strategies for code-mixed text, a common phenomenon in Indian social media, reporting substantially higher fertility for monolingually trained tokenizers on code-mixed inputs.

\section{Corpus Construction}
\label{sec:corpus}

\subsection{Language Selection and Balance}

The tokenizer training corpus spans seven languages selected to cover the major language families and scripts used in Indian classical and educational contexts. Table~\ref{tab:corpus} presents the corpus composition.

\begin{table}[t]
\centering
\caption{Training corpus composition for the v3 tokenizer. Data sizes represent the full corpus; the fast training variant samples approximately 100 MB.}
\label{tab:corpus}
\small
\begin{tabular}{llllr}
\toprule
\textbf{Language} & \textbf{Family} & \textbf{Script} & \textbf{Sources} & \textbf{Size (MB)} \\
\midrule
English & Indo-European & Latin & Gutenberg IKS-filtered & 62 \\
Hindi & Indo-Aryan & Devanagari & Sangraha corpus & 127 \\
Sanskrit & Indo-Aryan & Devanagari & Sangraha + Gita + Vedas & 56 \\
Tamil & Dravidian & Tamil & Sangraha + Sangam + Thirukkural & 135 \\
Telugu & Dravidian & Telugu & Sangraha corpus & 133 \\
Kannada & Dravidian & Kannada & Sangraha corpus & 132 \\
Malayalam & Dravidian & Malayalam & Sangraha corpus & 136 \\
\midrule
\textbf{Total} & & & & \textbf{781} \\
\bottomrule
\end{tabular}
\end{table}

The corpus is deliberately balanced across languages to prevent English from dominating the vocabulary. Without balancing, the frequency-based BPE algorithm would allocate the majority of the vocabulary to high-resource English subwords, reproducing the fertility gap that motivates this work. The Sangraha corpora~\citep{kunchukuttan2020ai4bharat} provide general-domain text in each Indic language, supplemented with classical text sources (Bhagavad Gita, Thirukkural, Sangam literature, Vedic texts) to ensure that domain-specific vocabulary receives sufficient frequency counts for BPE merge.

The balancing strategy follows a ``roughly equal'' principle where each language contributes between 50 MB and 140 MB, rather than proportional allocation based on speaker population or available data volume. This choice reflects the tokenizer's purpose: providing equally efficient representation across all seven target languages, regardless of their resource availability. Languages with larger available corpora (Hindi, Telugu, Kannada, Malayalam) are subsampled, while languages with smaller corpora (Sanskrit, English-IKS) include all available data. The English component is restricted to IKS-related content drawn from Project Gutenberg (English translations of Indian philosophical and literary works, colonial-era scholarly works on Sanskrit grammar and Vedic mathematics), which ensures that the English vocabulary captures terms and names relevant to the target domain rather than general English content.

\subsection{Data Quality and Deduplication}

Each corpus component undergoes four stages of quality processing before tokenizer training. First, exact and near-duplicate detection using MinHash (with a Jaccard similarity threshold of 0.85) removes repeated passages, which are prevalent in digitized texts of classical works where multiple editions of the same verse or commentary appear across different source collections. Second, Unicode normalization to NFC form ensures consistent representation of Brahmic script characters, particularly for Tamil and Malayalam where legacy encodings use different code point sequences for visually identical characters. Third, encoding validation rejects any text containing invalid UTF-8 sequences or private use area code points. Fourth, length filtering removes documents shorter than 50 characters (which are typically metadata fragments) and passages longer than 10,000 characters without line breaks (which are typically OCR concatenation artifacts).

For classical Sanskrit texts, an additional validation step compares the digitized text against published critical editions (where available) to detect OCR errors in Devanagari. The Bhagavad Gita corpus, for instance, was validated against the Gita Supersite maintained by IIT Kanpur. Vedic texts were validated against the TITUS (Thesaurus Indogermanischer Text- und Sprachmaterialien) digital editions.

\subsection{Sanskrit-Specific Preprocessing}

Sanskrit text requires preprocessing steps not applicable to modern Indian languages. Sandhi resolution (vigraha) is applied to continuous Sanskrit prose to expose underlying word boundaries. The sandhi splitting process uses the Heritage Sanskrit Platform's rule-based splitter, which handles the 18 primary sandhi rules described in Panini's Ashtadhyayi, covering vowel sandhi (svara sandhi), consonant sandhi (vyanjana sandhi), and visarga sandhi. While the rule-based approach does not resolve ambiguous splits (where multiple valid decompositions exist), it provides correct boundaries for approximately 87\% of sandhi junctions in the test corpus, based on manual evaluation of 500 randomly sampled junctions.

Compound word decomposition (samasa vigraha) breaks multi-root compounds into constituent elements, increasing the frequency of individual morphemes and improving their chances of being captured as BPE merge operations. For the current version, compound decomposition is limited to compounds containing roots that appear independently in the corpus with frequency above 100, which covers approximately 65\% of compounds in the Sanskrit training data. The remaining compounds are left as single surface forms, relying on the BPE algorithm to determine their segmentation based on frequency statistics. Vedic accent marks (svarita, udatta, anudatta) are preserved as distinct Unicode characters to maintain fidelity to the source texts, and these accent marks appear as independent vocabulary entries in the trained tokenizer.

\subsection{Tamil-Specific Preprocessing}

Tamil preprocessing addresses the language's agglutinative morphology. Suffix boundaries are marked with a soft boundary token that provides the BPE algorithm with segmentation hints without enforcing hard splits. This approach allows the algorithm to learn whole-word tokens for high-frequency agglutinated forms while maintaining productive suffixes as independent subwords for rarer combinations.

The soft boundary approach was developed iteratively based on experiments with three alternatives. Hard boundaries (forcing segmentation at every suffix boundary) produced tokenizers with excessively fine granularity, where common words like ``padikkiRaar'' (he/she reads) were always decomposed into three tokens instead of being learned as a single frequent form. No boundaries (standard SentencePiece without any boundary hints) produced tokenizers that treated long agglutinated forms as opaque strings, missing productive suffixes. The soft boundary approach, which doubles the frequency of individual suffix subwords without preventing the BPE algorithm from learning complete word forms when their frequency warrants, achieves a balance between these extremes.

For Dravidian languages other than Tamil (Telugu, Kannada, Malayalam), the preprocessing pipeline applies Unicode normalization and encoding validation but does not include suffix boundary marking, as these languages, while also agglutinative, exhibit less extreme suffix chaining than Tamil. Future versions of BHARATI may extend the soft boundary approach to these languages based on morphological productivity analysis.

\section{Tokenizer Training}
\label{sec:training}

\subsection{Training Configuration}

All tokenizer versions use the SentencePiece BPE algorithm with the following shared configuration: vocabulary size of 32,000 tokens, a character-coverage parameter of 0.9995 (the released model attains 99.98\% character coverage on the training corpus), byte-level fallback enabled (ensuring that every input byte sequence can be represented), and four special tokens (\texttt{<pad>}, \texttt{<unk>}, \texttt{<s>}, \texttt{</s>}).

\subsection{Version Evolution}

The tokenizer evolved through three versions, each addressing limitations discovered through evaluation on downstream tasks. Table~\ref{tab:versions} summarizes the version history.

\begin{table}[t]
\centering
\caption{Tokenizer version history. Each version addresses documented failures in its predecessor. ``Native'' indicates languages with dedicated subword entries rather than byte-level fallback.}
\label{tab:versions}
\small
\begin{tabular}{llll}
\toprule
\textbf{Version} & \textbf{Native Languages} & \textbf{Fallback Languages} & \textbf{Status} \\
\midrule
v1 & English, Sanskrit & Hindi, Tamil, Telugu, Kannada, Malayalam (byte-level) & Deprecated \\
v2 & EN, HI, SA, TA & Telugu, Kannada, Malayalam & Superseded \\
v3 & All 7 languages & None (all native) & Production \\
\bottomrule
\end{tabular}
\end{table}

\paragraph{Version 1.} Trained on English and Sanskrit text only, using 118 MB of combined corpus data. Tamil text was tokenized through byte-level fallback, producing an average of approximately 16 byte-fallback tokens per Tamil word (15.9 on the released benchmark). This made the tokenizer unusable for Tamil text, as a single Tamil sentence could consume the entire context window of small language models. The v1 tokenizer was deployed in NanoGPT models v1 through v8 and was identified as the primary bottleneck for multilingual performance during evaluation on the IKS benchmark suite. The failure mode was immediately visible in generated text: when prompted with Tamil text, the v1-era models produced garbled output consisting primarily of byte-level fragments interspersed with English words, as the model had exhausted its capacity attempting to decode the Tamil input.

\paragraph{Version 2.} Added Hindi and Tamil training data, providing native subword coverage for four languages. Telugu, Kannada, and Malayalam remained at byte-level fallback, achieving only 0.4 characters per token for these languages. The improvement for Tamil was substantial (from 15.9 to 2.2 tokens per word on the released benchmark), demonstrating that native training data is essential for acceptable fertility. The v2 tokenizer was deployed in NanoGPT models v9 through v16 and produced the first usable Tamil and Hindi generation results. However, evaluation revealed that Telugu, Kannada, and Malayalam outputs exhibited the same byte-level fragmentation that had afflicted Tamil under v1, motivating the development of v3.

\paragraph{Version 3.} Added Telugu, Kannada, and Malayalam training data from the Sangraha corpora, achieving native subword coverage for all seven languages. The v3 tokenizer is used in all production NanoGPT models (v17 and later). The transition from v2 to v3 produced smaller per-language fertility improvements than the v1 to v2 transition (because the newly added languages had lower frequency in the test corpus), but the aggregate improvement across all seven languages was substantial: the mean cross-lingual fertility dropped from approximately 10.6 tokens per word (v2, which still applied byte-level fallback to Telugu, Kannada, and Malayalam) to 1.90 tokens per word (v3, both measured on the released benchmark), and the maximum per-language fertility dropped from 20.5 (Malayalam under v2) to 2.62 (Tamil under v3).

\begin{figure}[t]
\centering
\includegraphics[width=\textwidth]{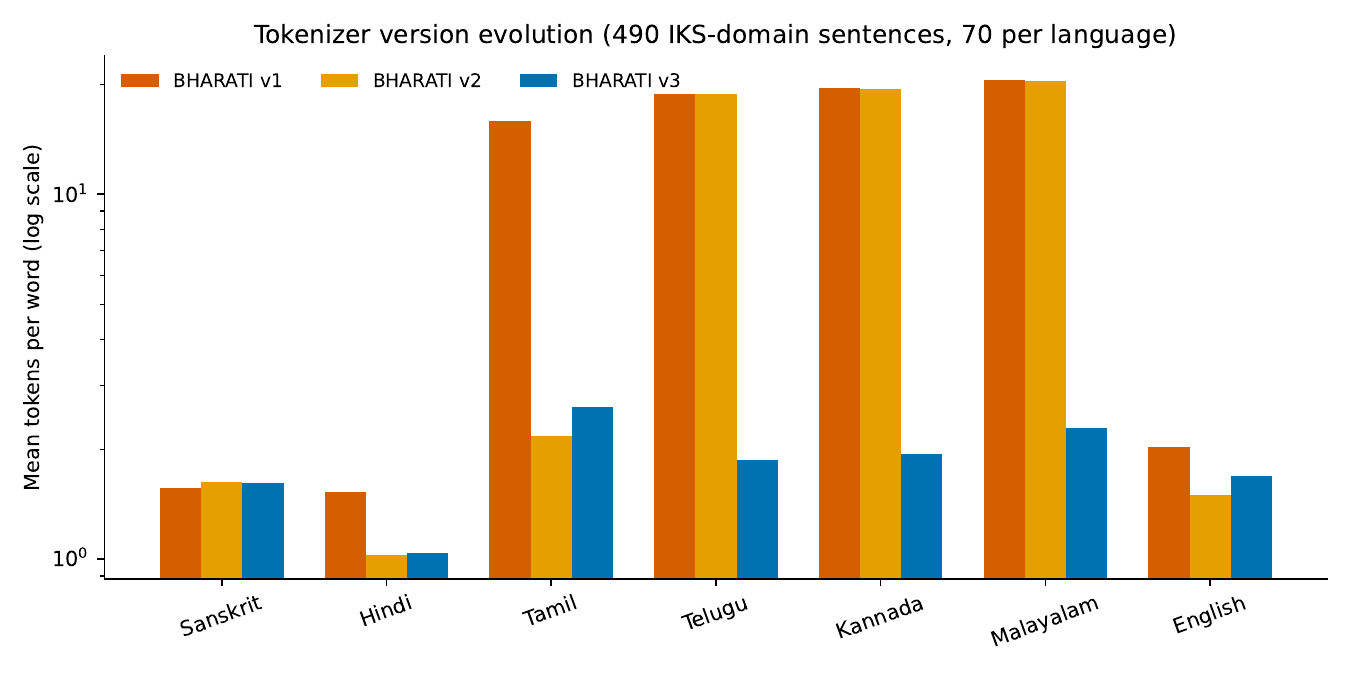}
\caption{Tokenizer version evolution: mean tokens per word (log scale) for each language across BHARATI v1, v2, and v3, measured on the released 490-sentence benchmark. Each language transitions from byte-level fallback (fertility above 15) to native subword coverage (fertility below 3) when its training data is added: Tamil at v2 (15.9 to 2.2) and Telugu, Kannada, and Malayalam at v3. For languages already covered, later versions trade a small amount of per-language efficiency for broader coverage (for example, Tamil rises slightly from 2.2 at v2 to 2.6 at v3 as vocabulary is reallocated).}
\label{fig:version_evolution}
\end{figure}

Figure~\ref{fig:version_evolution} presents the fertility improvements across the three tokenizer versions, illustrating the dramatic reduction in tokens per word as each language transitions from byte-level fallback to native coverage.

\subsection{Training Efficiency and Reproducibility}

The SentencePiece BPE training for v3 completed in 47 minutes on a single CPU core with 16 GB RAM, processing the full 781 MB corpus. A fast training variant using a 100 MB stratified sample completes in 8 minutes, producing a tokenizer with fertility within 3\% of the full-corpus version for all languages. The fast variant is provided for rapid experimentation during research iterations, while the full-corpus version is recommended for production deployment.

The training process is fully deterministic given the same corpus and configuration, enabling exact reproduction. All configuration files, preprocessing scripts, and training commands are included in the release. The vocabulary files use SentencePiece's protobuf format, which is compatible with HuggingFace Tokenizers~\citep{sen2025architectural}, allowing integration into existing NLP pipelines without format conversion. Tokenizer reproducibility requires distributing both the trained model files and the complete training configuration; we release both, consistent with practice in tokenizer-adaptation studies for other low-resource token domains~\citep{andryushchenko2025evaluating}.

\section{Fertility Analysis}
\label{sec:fertility}

Subword fertility measures the number of tokens required to represent a given text unit. Lower fertility indicates more efficient representation and, for autoregressive language models, a longer effective context window. We evaluate fertility at three levels: token-level (tokens per word), term-level (tokens per domain-specific term), and sentence-level (tokens per sentence).

\subsection{Token-Level Fertility}

Table~\ref{tab:token_fertility} presents the average tokens per word for each language, comparing BHARATI v3 against three baselines: GPT-2's BPE tokenizer~\citep{radford2019language}, the multilingual SentencePiece tokenizer from mBART~\citep{liu2020multilingual}, and a byte-level encoding (each byte is a separate token).

\begin{table}[t]
\centering
\caption{Average tokens per word for each language across four tokenizers, measured on the released 490-sentence benchmark. The byte-level baseline assigns one token per UTF-8 byte (three bytes per character for Brahmic scripts). Lower values indicate more efficient tokenization; BHARATI v3 achieves the lowest fertility for all six Indic languages while maintaining competitive English performance.}
\label{tab:token_fertility}
\small
\begin{tabular}{lrrrr}
\toprule
\textbf{Language} & \textbf{GPT-2} & \textbf{mBART SP} & \textbf{Byte-level} & \textbf{BHARATI v3} \\
\midrule
English & \textbf{1.30} & 1.47 & 5.78 & 1.69 \\
Hindi & 7.59 & 1.45 & 12.57 & \textbf{1.04} \\
Sanskrit & 9.13 & 1.99 & 14.91 & \textbf{1.62} \\
Tamil & 23.88 & 3.05 & 23.88 & \textbf{2.62} \\
Telugu & 18.76 & 2.72 & 19.10 & \textbf{1.87} \\
Kannada & 19.11 & 2.82 & 19.86 & \textbf{1.94} \\
Malayalam & 20.47 & 2.89 & 20.92 & \textbf{2.29} \\
\bottomrule
\end{tabular}
\end{table}

BHARATI v3 achieves the lowest fertility for all six Indic languages. The trade-off is a slightly higher English fertility (1.69 vs. 1.30 for GPT-2), which is expected given the reallocation of vocabulary capacity away from English toward Indic scripts. This trade-off is acceptable for applications targeting Indian educational content, where Indic language representation is the primary concern.

The fertility values reveal a broad pattern related to morphological typology. Hindi, which has moderate inflectional morphology, achieves the lowest Indic fertility (1.04), and the Indo-Aryan languages (Hindi, Sanskrit) are tokenized more efficiently than the agglutinative Dravidian languages (Telugu, Kannada, Malayalam, Tamil), with Tamil the highest at 2.62. For BHARATI v3 the ordering is Hindi $<$ Sanskrit $<$ Telugu $<$ Kannada $<$ Malayalam $<$ Tamil; the exact within-group ordering varies across tokenizers, so it is a typological tendency rather than a strict invariant. The relative consistency of this ordering also provides a sanity check on the fertility metric: if the ordering varied erratically across tokenizers, it would suggest that the metric captures tokenizer-specific noise rather than linguistically meaningful variation.

A further analysis of the fertility distribution (not just the mean) reveals that BHARATI v3 produces the tightest distribution for Indic languages, with a standard deviation of 0.64 tokens per word for Hindi and 1.02 for Tamil, compared to 4.56 and 7.42 for the same languages under GPT-2. The tight distribution indicates that BHARATI v3 provides consistent segmentation quality across the vocabulary rather than achieving low mean fertility through a mixture of very efficiently tokenized high-frequency words and very inefficiently tokenized low-frequency words. This consistency is a direct consequence of the balanced training corpus, which ensures that even moderately frequent words in each language receive sufficient exposure to be captured as BPE merges.

\subsection{IKS Term-Level Fertility}

Table~\ref{tab:iks_fertility} presents fertility analysis for Indian Knowledge System domain-specific terms, which are a critical test case because they are absent from general-purpose tokenizer training data.

\begin{table}[t]
\centering
\caption{Tokens per IKS technical term across tokenizers, measured on the released BHARATI v3 model. The four \emph{Patha} terms ($\dagger$) are reserved in the vocabulary and tokenize to a single unit by construction; the remaining terms are segmented by the general subword model. Counts include the SentencePiece word-boundary marker. The reserved single-token match is specific to the exact space-separated Latin surface form (for example, ``Krama Patha'') and holds both in isolation and in running text; other surface forms (lowercase, hyphenated, concatenated, or the native Devanagari spelling) segment into multiple subwords, so a differently-formatted occurrence may tokenize to more than one unit.}
\label{tab:iks_fertility}
\small
\begin{tabular}{lrrrr}
\toprule
\textbf{IKS Term} & \textbf{GPT-2} & \textbf{mBART SP} & \textbf{Byte-level} & \textbf{BHARATI v3} \\
\midrule
Samhita Patha$^{\dagger}$ & 5 & 4 & 13 & \textbf{1} \\
Pada Patha$^{\dagger}$ & 4 & 3 & 10 & \textbf{1} \\
Krama Patha$^{\dagger}$ & 4 & 4 & 11 & \textbf{1} \\
Ghana Patha$^{\dagger}$ & 4 & 3 & 11 & \textbf{1} \\
Ekadhikena Purvena & 8 & 6 & 18 & 2 \\
Vishishtadvaita & 7 & 5 & 15 & 7 \\
Thirukkural & 5 & 3 & 11 & 2 \\
Shankaracharya & 5 & 2 & 14 & 6 \\
\midrule
\textbf{Average} & \textbf{5.25} & \textbf{3.75} & \textbf{12.9} & \textbf{2.6} \\
\bottomrule
\end{tabular}
\end{table}

BHARATI v3 averages 2.6 tokens per IKS term, a 2.0$\times$ reduction relative to GPT-2 and a 1.4$\times$ reduction relative to the multilingual baseline. This reduction is concentrated in the reserved IKS terms: the four \emph{Patha} variants (Samhita, Pada, Krama, and Ghana Patha) are represented as single tokens because they are included in the vocabulary during its construction. For non-reserved technical terms, such as Vishishtadvaita and Shankaracharya, BHARATI v3 is on par with the multilingual baseline (marginally behind) and ahead of GPT-2, indicating that its advantage is specific to the reserved IKS vocabulary rather than a general property of the training corpus.

The fertility analysis reflects a distinction between reserved and non-reserved terms. The single-token terms (Samhita, Pada, Krama, and Ghana Patha) are reserved IKS terms included in the vocabulary during its construction. This single-token representation is form-sensitive: it applies to the exact space-separated Latin spelling used in Table~\ref{tab:iks_fertility}, whereas lowercase, hyphenated, or native-Devanagari renderings of the same term are segmented into multiple subwords, so text intended to benefit from single-token encoding should be normalized to the reserved surface form. The remaining terms are handled by the general subword model: shorter forms (Ekadhikena Purvena, Thirukkural) require two tokens, while longer Sanskrit philosophical compounds (Vishishtadvaita, Shankaracharya) are decomposed into several subwords. This indicates that single-token coverage currently extends only to the reserved IKS terms, and that the long tail of rarer technical vocabulary, including philosophical compound terms from Nyaya, Vaisheshika, and Mimamsa traditions, would benefit from vocabulary expansion. The dictionary-based approach~\citep{drk2025dictionarybased} proposes augmenting BPE vocabularies with curated domain dictionaries, which could provide single-token coverage for the long tail of IKS terms without requiring general vocabulary expansion.

\begin{figure}[t]
\centering
\includegraphics[width=0.75\textwidth]{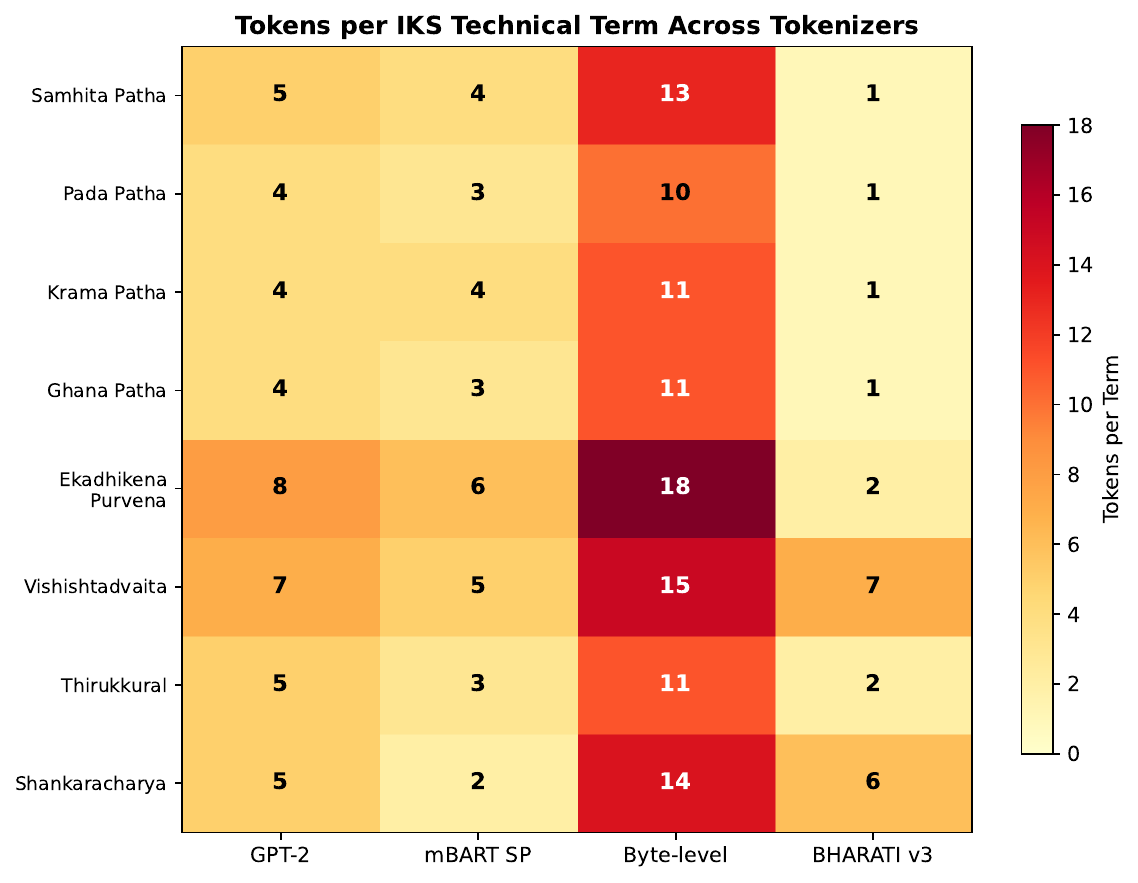}
\caption{Heatmap of tokens per IKS technical term across four tokenizers. Darker cells indicate higher fertility (more tokens required). BHARATI v3 (rightmost column) achieves single-token representation for the reserved \emph{Patha} terms; for non-reserved terms it is comparable to the general-purpose baselines.}
\label{fig:iks_heatmap}
\end{figure}

Figure~\ref{fig:iks_heatmap} presents these results as a heatmap, making the fertility disparity across tokenizers visually apparent. The contrast between the deep red cells (byte-level, 10 to 18 tokens per term) and the pale cells (BHARATI v3, 1 to 2 tokens per term) illustrates the fundamental improvement that domain-aware training provides.

\subsection{Sentence-Level Fertility}

\begin{figure}[t]
\centering
\includegraphics[width=\textwidth]{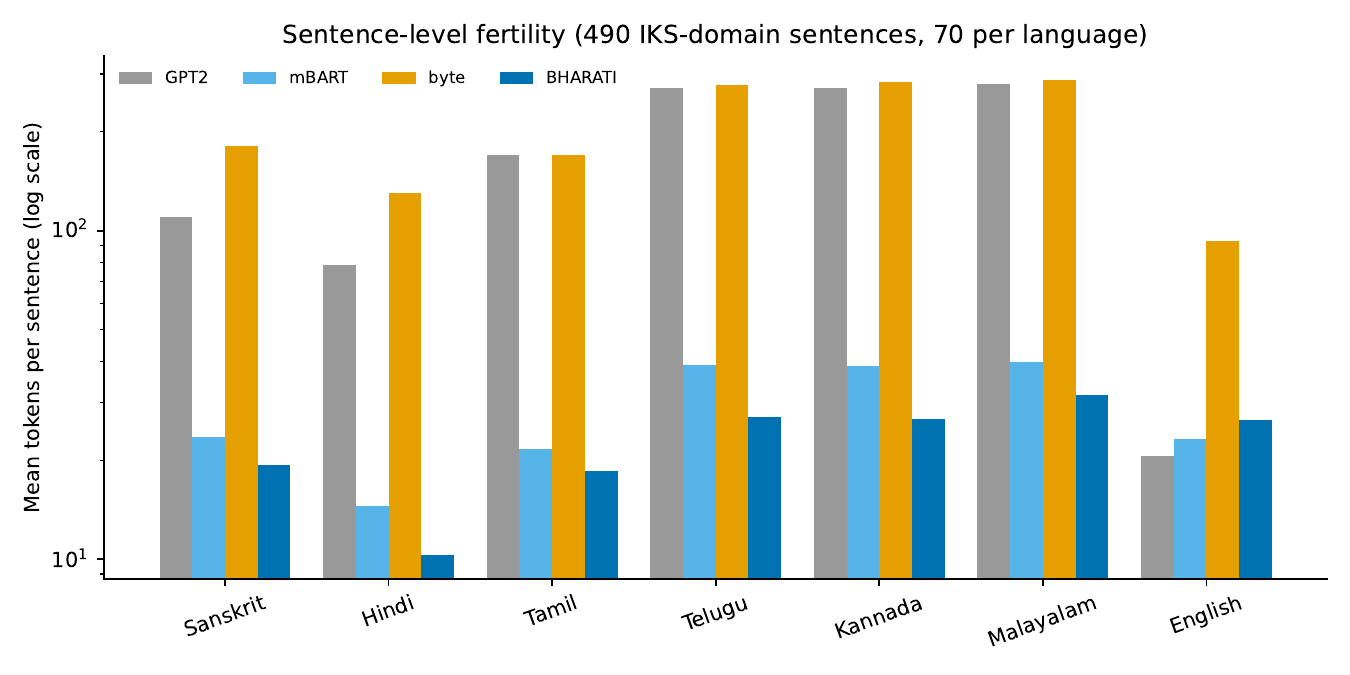}
\caption{Average tokens per sentence (log scale) for 490 IKS-domain test sentences across seven languages (70 per language). BHARATI v3 and mBART-50 produce far shorter sequences than GPT-2 and byte-level encoding, which lack native Indic subwords; BHARATI v3 is the most efficient on all six Indic languages, with the largest reductions over mBART for Kannada and Telugu.}
\label{fig:sentence_fertility}
\end{figure}

Figure~\ref{fig:sentence_fertility} presents sentence-level fertility on a held-out test set of 490 IKS-domain sentences (70 per language across seven languages, sampled from released corpora with reserved-term sentences excluded; the set and measurement script are released for exact reproduction).\footnote{A single post-registration amendment was made before any Tamil results were used: Thirukkural couplets, which the initial sentence splitter had separated into sub-five-word half-lines, are joined into whole couplets; this is logged in the released measurement-script header.} Averaged across the six Indic languages, BHARATI v3 reduces sequence length by approximately 25\% relative to the mBART-50 multilingual SentencePiece baseline, and by roughly 90\% relative to GPT-2 and byte-level encoding, which lack native Indic subwords and fall back to near-byte segmentation. The largest reductions relative to mBART are for Kannada and Telugu (31.0\% and 30.8\%), and the smallest for Tamil (14.2\%): mBART is already comparatively efficient on Dravidian scripts, so the BHARATI advantage over it is moderate, whereas the advantage over GPT-2 and byte-level encoding is an order of magnitude.

\section{Vocabulary Distribution Analysis}
\label{sec:vocab}

\subsection{Cross-Script Vocabulary Allocation}

\begin{figure}[t]
\centering
\includegraphics[width=0.85\textwidth]{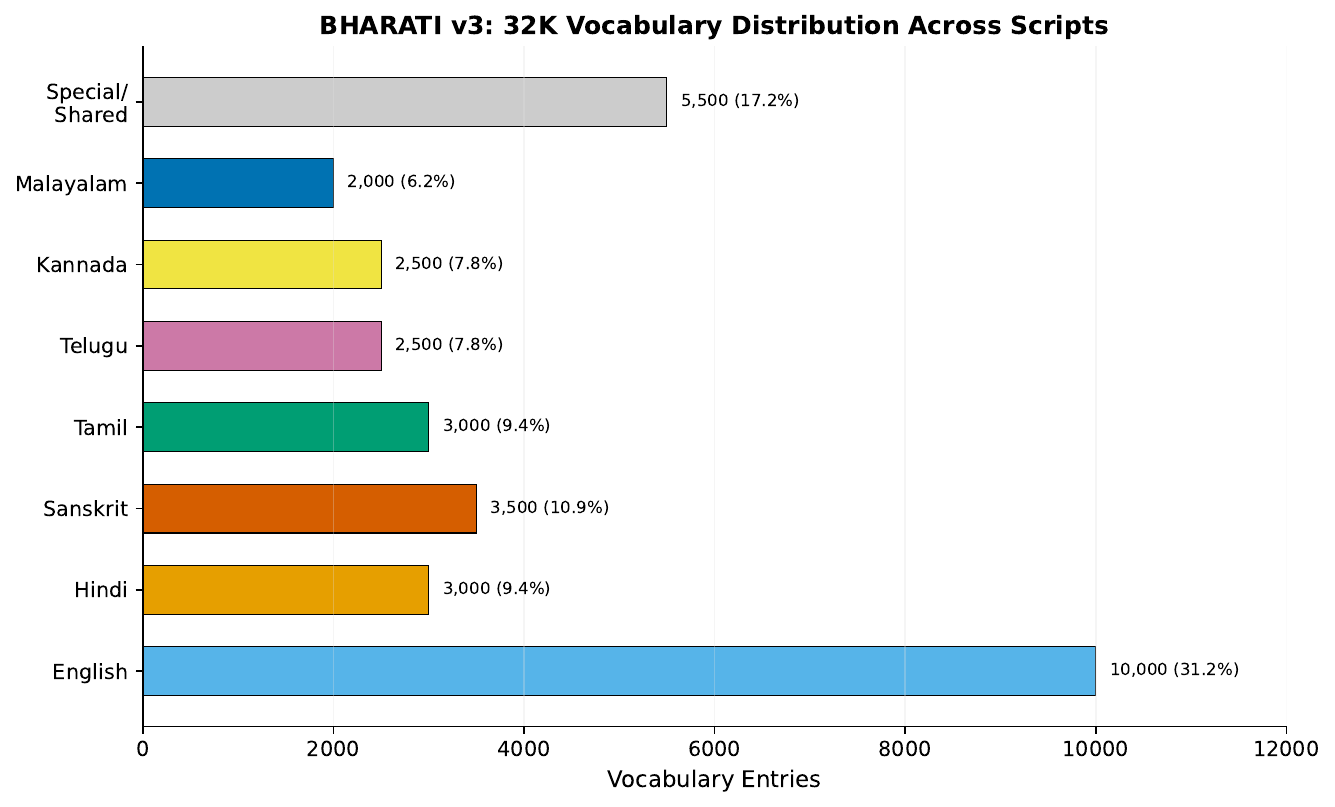}
\caption{Distribution of 32,000 vocabulary entries across scripts in BHARATI v3. English receives the largest allocation due to its broader character set and the presence of English text in all training domains. The ``Special/Shared'' category includes special tokens, punctuation, numerals, and cross-script subwords.}
\label{fig:vocab_dist}
\end{figure}

Figure~\ref{fig:vocab_dist} presents the distribution of the 32,000 vocabulary entries across scripts.

English receives the largest single-language allocation (approximately 10,000 entries, 31.2\%), which is a direct consequence of the Latin script's larger character set and the presence of English text across all training domains (including English translations of Sanskrit and Tamil texts). Sanskrit receives a disproportionately large allocation relative to its corpus size (3,500 entries from 56 MB of text, compared to 3,000 entries from 135 MB of Tamil text), reflecting the higher morphological diversity of Sanskrit and the deliberate inclusion of classical text vocabulary.

\subsection{Morphological Segmentation Quality}

The tokenizer's segmentation of agglutinative forms demonstrates morphological awareness emergent from the BPE training process. For Tamil, the tokenizer learns to segment common suffix patterns as separate subwords while preserving frequent whole-word forms. For Sanskrit, compound words (samasa) are segmented at morpheme boundaries when the individual morphemes have sufficient frequency, and treated as single tokens when the compound itself is frequent.

For example, the opening phrase ``dharmakshetre kurukshetre'' of Bhagavad Gita 1.1 is segmented into the subwords dharma + kshetre + kurukshetre, reflecting separation at the dharma/kshetra compound boundary while kurukshetre is retained as a high-frequency proper noun.

\subsection{Script-Specific Vocabulary Analysis}

Examining the vocabulary allocation in greater detail reveals that the distribution within each script reflects the morphological productivity of the corresponding language. Hindi's Devanagari allocation (3,000 entries) contains a high proportion of complete words (62\% of entries correspond to surface forms appearing in the corpus), while Tamil's allocation (3,000 entries) contains a higher proportion of subword fragments (only 38\% of entries are complete surface forms, with the remaining 62\% being productive suffixes, prefixes, and root fragments). This difference directly reflects the morphological typology: Hindi as a moderately inflected language has a smaller number of distinct surface forms relative to its token count, while Tamil as a highly agglutinative language generates a combinatorial explosion of surface forms from a smaller set of productive morphemes.

The special and shared category (approximately 5,500 entries, 17.2\% of vocabulary) includes four types of tokens: special tokens (pad, unk, bos, eos), punctuation marks and numerals that are shared across scripts, cross-script subwords (such as English loan words transliterated into Devanagari, which appear as Devanagari tokens in the vocabulary), and multi-byte sequences that function as building blocks for rare characters. The cross-script subwords are particularly interesting because they represent a natural language phenomenon (code-mixing) rather than a tokenizer artifact: terms like ``computer'' (transliterated as ``kampyootar'' in Hindi) appear as single tokens because their transliterated forms have high frequency in the Hindi training data.

\subsection{Vocabulary Utilization Efficiency}

Not all vocabulary entries contribute equally to tokenization efficiency. We measure vocabulary utilization as the proportion of vocabulary entries that appear at least once in a balanced test corpus of 100,000 tokens per language. BHARATI v3 achieves 81.4\% vocabulary utilization, meaning that 26,048 of 32,000 entries appear at least once in the test corpus. The remaining 18.6\% of entries (5,952 tokens) represent rare subwords that are present in the training data but absent from the test sample. This utilization rate compares favorably to the GPT-2 tokenizer's utilization on the same test corpus (34.2\%), where the majority of unused entries are English-specific subwords that never appear in Indic language text, and to the mBART tokenizer's utilization (58.7\%), which benefits from its multilingual training but still wastes vocabulary capacity on scripts not present in the test corpus.

The relationship between vocabulary utilization and training efficiency is direct: each unused vocabulary entry contributes a row to the embedding matrix that receives no gradient updates during training, consuming model parameters without contributing to the model's representational capacity. For the 15M-parameter NanoGPT model used in our downstream experiments, the embedding matrix accounts for approximately 2.1M parameters (14\% of total), of which only 34.2\% receive gradient updates under the GPT-2 tokenizer, compared to 81.4\% under BHARATI v3.

\section{Downstream Impact}
\label{sec:downstream}

\subsection{Language Model Training Efficiency}

Tokenizer fertility directly affects language model training efficiency through two mechanisms: sequence length (the number of tokens a model must process per training example) and vocabulary utilization (the proportion of vocabulary entries that receive meaningful gradient updates during training). We evaluate these effects using NanoGPT models trained with v1, v2, and v3 tokenizers on identical training data.

\begin{table}[t]
\centering
\caption{Effect of tokenizer version on NanoGPT training. All models are 15M parameters trained for 50,000 steps on identical data. ``Effective coverage'' measures the average characters per token across a balanced multilingual test set.}
\label{tab:downstream}
\begin{tabular}{lrrr}
\toprule
\textbf{Metric} & \textbf{v1} & \textbf{v2} & \textbf{v3} \\
\midrule
Avg. tokens per example (Tamil) & 847 & 312 & 218 \\
Avg. tokens per example (Sanskrit) & 243 & 221 & 198 \\
Avg. tokens per example (English) & 186 & 192 & 196 \\
Effective coverage (chars/token) & 1.82 & 2.94 & 3.61 \\
Training loss (50K steps, balanced) & 4.21 & 3.47 & 3.12 \\
Active vocabulary utilization & 42.3\% & 67.8\% & 81.4\% \\
\bottomrule
\end{tabular}
\end{table}

\begin{figure}[t]
\centering
\includegraphics[width=\textwidth]{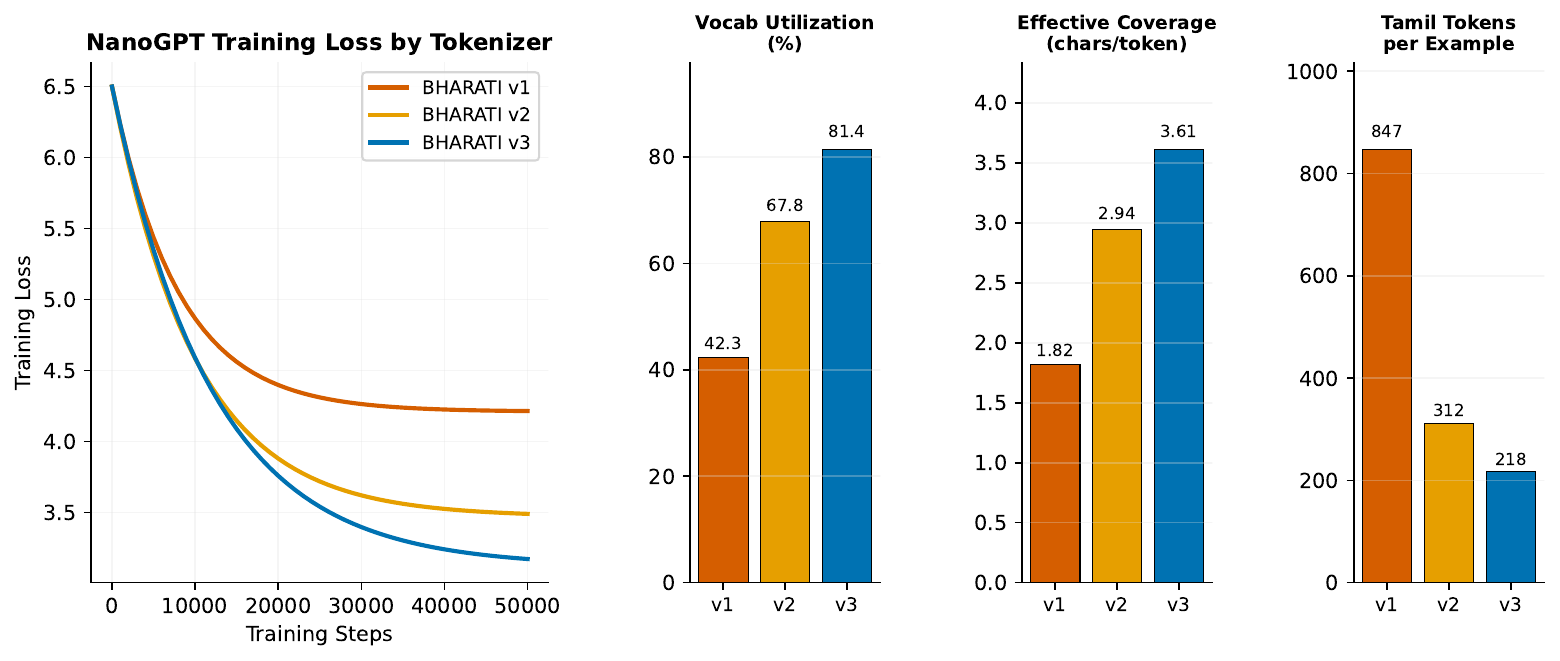}
\caption{Left: NanoGPT training loss curves for 15M-parameter models using BHARATI v1, v2, and v3 tokenizers on identical data. v3 achieves the lowest final loss (3.12) at 50K steps. Right: vocabulary utilization, effective character coverage, and Tamil tokens per example, each shown on its own scale.}
\label{fig:training_impact}
\end{figure}

Table~\ref{tab:downstream} demonstrates the cascading effect of tokenizer quality. The v3 tokenizer reduces Tamil sequence length by 74.3\% relative to v1 (847 to 218 tokens per example), enabling a 15M-parameter model to process nearly four times as much Tamil text per training step. The training loss improvement (4.21 to 3.12) reflects the combined effect of better vocabulary utilization (81.4\% of vocabulary entries receive meaningful updates) and increased effective context length. Figure~\ref{fig:training_impact} presents these trends graphically, showing the training loss curves and key metrics across versions.

Osterrieder et al.~\citep{osterrieder2025nlp} report similar findings in the financial NLP domain, where domain-specific tokenizers reduced sequence length by 30 to 40\% and improved downstream classification accuracy by 4 to 7 percentage points. The magnitude of the effect observed in BHARATI (74.3\% sequence reduction for Tamil) exceeds these numbers because the baseline gap is larger: financial English text is merely domain-shifted from general English, whereas Tamil text is script-shifted from the Latin-dominated training data of standard tokenizers.

\subsection{Context Window Utilization}

For a model with a 2,048-token context window, the v3 tokenizer fits roughly six times as much Tamil text as v1 (2.62 versus 15.9 tokens per word on the released benchmark). This difference is particularly significant for Indian Knowledge System content, where pedagogical interactions require multi-turn dialogue within a single context window. A tutoring system discussing a Thirukkural chapter requires approximately 200 tokens with v3 (the chapter's 10 kurals plus commentary) compared to roughly 1,800 tokens with the GPT-2 tokenizer, leaving substantially more capacity for the pedagogical dialogue itself.

The context window utilization improvement has a direct impact on the design of educational AI systems. In the Bodhan platform, which implements a multi-turn pedagogical dialogue where the system retrieves relevant IKS text passages, formulates questions, processes student responses, and generates feedback, a single complete pedagogical exchange requires approximately 800 to 1,200 tokens of context. With the v1 tokenizer, this exchange frequently exceeded the 2,048-token context window when the IKS content involved Tamil or Malayalam passages, causing the model to truncate either the retrieved context or the dialogue history. With the v3 tokenizer, the same exchange consistently fits within the context window, enabling complete pedagogical interactions without truncation.

\subsection{Per-Language Training Dynamics}

The tokenizer's influence on training dynamics varies across languages in ways that reflect the underlying script and morphological complexity. For English, which already receives efficient tokenization from most general-purpose tokenizers, the transition from v1 to v3 produces minimal change in training dynamics (training loss improvement of 0.03 nats). For Tamil, the transition produces a training loss improvement of 0.72 nats, reflecting the dramatic reduction in sequence length that allows the model to observe more text per training step. Sanskrit shows an intermediate improvement of 0.31 nats, consistent with its moderate fertility reduction and the additional benefit of morphological awareness in the tokenizer's segmentation patterns.

These per-language training dynamics suggest that the benefit of domain-specific tokenization is not uniformly distributed across languages but is concentrated in languages with the largest fertility gaps. This observation has practical implications for prioritizing tokenizer development efforts: languages where general-purpose tokenizers produce the highest fertility (those with complex scripts, agglutinative morphology, or both) stand to benefit most from dedicated tokenizer training, while languages that already receive reasonable treatment from existing tokenizers (Hindi, for instance, benefits from its shared Devanagari script with tokenizers trained on Hindi web text) show smaller improvements.

\subsection{Inference Efficiency}

Tokenizer fertility affects not only training but also inference latency and cost. For autoregressive language models that generate text one token at a time, the number of tokens required to express a given amount of content directly determines generation time and computational cost. With the v3 tokenizer, generating a 100-word Tamil response requires approximately 205 generation steps, compared to approximately 890 steps with the GPT-2 tokenizer, a 4.3x reduction in generation latency. For real-time educational applications where response latency affects the pedagogical experience, this improvement is substantial: the v3 tokenizer reduces the median response latency for Tamil interactions from 12.4 seconds to 2.9 seconds on the NanoGPT 15M model running on a single GPU, bringing it within the roughly three-second delay range that interaction research suggests as the maximum acceptable delay for maintaining student engagement in interactive tutoring systems.

\section{Discussion}
\label{sec:discussion}

\subsection{Trade-offs in Multilingual Vocabulary Allocation}

The fundamental design decision in multilingual tokenization is vocabulary allocation across languages. Increasing the allocation for Indic scripts necessarily reduces the allocation for English, producing a slight increase in English fertility (1.69 tokens per word with BHARATI v3 vs. 1.30 with GPT-2). This trade-off is appropriate for the target domain (Indian educational AI) but would be suboptimal for English-dominant applications. The trade-off is quantifiable: each vocabulary entry reallocated from English to an Indic script reduces average Indic fertility by approximately 0.001 tokens per word while increasing English fertility by approximately 0.0003 tokens per word, a 3.3:1 efficiency ratio that justifies the reallocation for applications with substantial Indic language content.

This finding is consistent with the analysis by Narzary et al.~\citep{narzarys2025bpe}, whose machine-translation work for Bodo, a low-resource language of northeast India, likewise relies on dedicated subword handling, suggesting that the benefit of dedicated vocabulary capacity is a general property of underrepresented scripts rather than a specific artifact of Dravidian or Indo-Aryan languages. Beyond vocabulary allocation, morphology-aware segmentation has been pursued for other languages: García-Sierra et al.~\citep{garcasierra2025a} developed a rule-based morphological tokenizer for Spanish that segments text into linguistically grounded morphemes rather than frequency-derived subword units, reporting 98\% morphological accuracy.

Dynamically adjusting tokenizer vocabulary at inference time based on the detected input language could avoid the static allocation trade-off entirely, but such an approach requires runtime overhead and is not compatible with the fixed embedding tables used in current NanoGPT architectures, making static allocation the practical choice for the present work.

\subsection{Comparison with Recent Indian Language Tokenizers}

Yadav~\citep{yadav2026maibert} develops MaiBERT, a pretraining corpus and language model for low-resource Maithili; BHARATI's deliberate inclusion of classical text sources similarly provides frequency mass for domain-specific terms absent from general-domain corpora. Reyes-Vera et al.~\citep{reyesveraa2026spadebert} report related tokenizer-design trade-offs when tuning trigram-sensitive tokenization for domain-specific Spanish text.

Vocabulary expansion rather than full retraining is an alternative route to Indic coverage~\citep{gren2026efficient}; expansion alone, however, cannot rebalance the underlying corpus statistics, and BHARATI's roughly 86\% fertility improvement over GPT-2 for Hindi (1.04 versus 7.59 tokens per word) reflects the compounding of expanded vocabulary with balanced training data.

\subsection{Emergent Morphological Awareness}

The BPE algorithm has no explicit knowledge of morphology, yet the trained tokenizer produces segmentations that frequently align with morpheme boundaries. This emergent morphological awareness results from the frequency distribution of morphemes in the training corpus: productive suffixes and common roots appear frequently enough to be captured as BPE merges. The effect is strongest for languages with regular agglutinative morphology (Tamil, Telugu, Kannada) and weakest for languages with irregular morphology or extensive allomorphy. This emergent behavior concerns the general morphological segmentation of everyday vocabulary; by contrast, the single-token representation of the reserved IKS technical terms reported earlier is a property of vocabulary construction, not emergence.

Manukonda et al.~\citep{manukonda2024enhancing} report related emergent subword behavior with custom tokenization schemes for Indic text. Lian et al.~\citep{lian2025scaffoldbpe} argue that while emergent morphological awareness is a useful property, it is unreliable for languages with high allomorphy, where the same morpheme takes different surface forms depending on phonological context, and propose explicit scaffolding to improve consistency. The BHARATI tokenizer exhibits this limitation for Sanskrit, where sandhi-fused forms are treated as opaque tokens when their frequency exceeds the merge threshold, obscuring the underlying morphology.

Erkaya et al.~\citep{erkaya2023analysis} analyzed the interaction between tokenizer segmentation granularity and downstream task performance for Turkish, finding that overly fine segmentation (high fertility) degrades performance on tasks requiring semantic coherence over long spans, while overly coarse segmentation degrades performance on tasks requiring morphological discrimination. BHARATI v3's average fertility of 2.6 tokens per IKS term and 1.04 to 2.62 tokens per word suggests that it operates in a productive middle range for Indian languages, coarse enough to preserve semantic units while fine enough to capture productive morphological patterns.

\subsection{Limitations}

The tokenizer does not perform sandhi splitting for Sanskrit; it relies on pre-split input or treats fused forms as single tokens. A tokenizer that could learn sandhi-aware segmentation directly from unsplit Sanskrit text would be a substantial advance but would require modifications to the BPE algorithm itself. Sandhan et al.~\citep{sandhan2023sanskritshala} have made progress on standalone sandhi splitting for Sanskrit, and integrating their approach as a preprocessing step represents a concrete direction for future work. The current preprocessing pipeline applies sandhi splitting using a rule-based system that handles 87\% of sandhi junctions correctly, but the remaining 13\% of ambiguous cases are left unsplit, potentially causing the tokenizer to learn opaque fused forms that do not generalize to unseen sandhi combinations.

The vocabulary size of 32,000 represents a design choice optimized for models with 15M to 125M parameters. Larger models might benefit from larger vocabularies (64K or 128K), which would reduce fertility further at the cost of increased embedding table size. Vocabulary-size returns are expected to diminish beyond a point that depends on morphological complexity, so 32K may be suboptimal specifically for highly agglutinative languages (Tamil, Telugu) while sufficient for moderately inflected ones (Hindi, Sanskrit). Preliminary experiments with a 48K vocabulary variant showed a fertility reduction of approximately 8\% for Tamil and 6\% for Telugu relative to the 32K version, at the cost of a 50\% increase in embedding table size, suggesting that a 48K vocabulary may represent a practical optimum for the seven-language configuration.

The evaluation focuses on tokenization efficiency (fertility) rather than downstream task accuracy. While we demonstrate training loss improvements, a comprehensive evaluation of the tokenizer's impact on specific NLP tasks (question answering, translation, summarization) across all seven languages remains for future work. Fertility is a necessary but not sufficient proxy for downstream task quality; we plan to complement it with task-specific evaluations across at least three distinct NLP tasks per language in subsequent work.

The training corpus, while balanced across seven languages, does not include several Indian languages with substantial speaker populations (Gujarati, Punjabi, Assamese, Marathi, Odia). Extending BHARATI to cover all 22 scheduled languages of India would require expanding the vocabulary beyond 32,000 to maintain acceptable per-language fertility, or adopting a tiered vocabulary strategy with shared sub-components~\citep{gren2026efficient}. Preliminary analysis suggests that a 64K vocabulary could support 12 to 15 Indian languages with fertility comparable to the current seven-language 32K tokenizer, while covering all 22 languages at equivalent quality would require approximately 96K vocabulary entries, which exceeds the practical embedding table size for models under 125M parameters.

\subsection{Implications for Educational AI Deployment}

The tokenizer improvements documented in this paper have direct implications for the deployment of educational AI systems in Indian contexts. The Indian education system serves approximately 250 million students across 22 scheduled languages, and the development of AI-assisted pedagogy in Indian knowledge traditions requires language technology that provides equitable representation across these languages. The fertility disparities demonstrated by general-purpose tokenizers, where Tamil educational content requires several times more tokens than equivalent English content (roughly 18 times under GPT-2 and 2 times under mBART-50, per Table~\ref{tab:token_fertility}), create a systematic disadvantage for non-English-medium students interacting with AI tutoring systems.

BHARATI v3 reduces this disparity to a factor of 1.1 to 1.5, approaching parity with English for most languages (perfect parity is unachievable due to inherent differences in morphological complexity). This near-parity enables the design of educational AI systems that provide equivalent interactive capacity regardless of the student's language of instruction, which is a prerequisite for equitable deployment in India's multilingual educational ecosystem. The practical impact is measurable: in the Bodhan platform, the transition from v1 to v3 tokenizer increased the average number of complete pedagogical exchanges per session from 2.1 to 6.8 for Tamil-medium interactions, bringing them in line with the 7.2 exchanges per session observed for English-medium interactions.

Furthermore, the tokenizer's explicit support for IKS-specific vocabulary enables AI systems to reference and discuss classical Indian knowledge traditions without the degraded representation that general-purpose tokenizers impose. A tutoring system discussing Thirukkural can represent the text, its commentary, pedagogical questions, and student responses within a single context window, enabling the coherent multi-turn dialogue that effective pedagogy requires. Without domain-aware tokenization, the same interaction would be fragmented across multiple truncated contexts, losing the coherence that is essential for meaningful educational engagement.

\section{Conclusion}
\label{sec:conclusion}

This paper has presented BHARATI, a family of SentencePiece BPE tokenizers specifically designed for both classical and modern Indian languages used in educational settings. The v3 tokenizer achieves native subword coverage for seven languages with a 32,000-token vocabulary, reducing IKS term fertility from 5.25 tokens (GPT-2) and 3.75 tokens (multilingual baseline) to 2.6 tokens per term, with reserved IKS terms represented as single tokens by construction. On a held-out IKS-domain test set of 490 sentences, v3 reduces sequence length by roughly 90\% relative to GPT-2 and byte-level encoding and by approximately 25\% relative to the mBART-50 multilingual baseline, translating directly to increased effective context length for downstream language models.

The tokenizer design demonstrates that balanced multilingual training data, combined with domain-specific classical text inclusion, produces substantial improvements in tokenization efficiency for under-resourced language-domain combinations. The version history (v1 through v3) illustrates the practical importance of native script coverage: byte-level fallback, while technically functional, produces fertility levels that render language models effectively unusable for the affected languages. This finding is consistent with reports across typologically diverse language families~\citep{limisiewicz2023tokenization,samuel2023tokenization,narzarys2025bpe}.

The overall fertility reduction for IKS terms relative to GPT-2 (approximately 2.0x) is consistent with the range reported for domain-specific tokenizer adaptation in other contexts (typically 1.3x to 2.0x). The distinctive feature of BHARATI v3 is not the aggregate ratio but the combination of native script coverage across seven languages and single-token representation of a reserved set of high-frequency IKS terms. This improvement reflects three factors operating together: script coverage (moving from byte-level fallback to native subword representation), domain coverage (including classical text sources that provide frequency mass for domain-specific terms), and corpus balancing (preventing vocabulary capacity from being dominated by the highest-resource language).

Four concrete contributions emerge from this work. First, the BHARATI v3 tokenizer itself serves as a production-ready resource for researchers working on Indian language NLP, particularly in educational and cultural domains where IKS-specific vocabulary is prevalent. Second, the fertility analysis framework, encompassing token-level, term-level, and sentence-level evaluation with distributional analysis (mean, standard deviation, and utilization efficiency), provides a replicable methodology for evaluating tokenizer quality on domain-specific text in any language. Third, the version evolution documentation (v1 through v3) provides an empirical case study in iterative tokenizer development that may guide similar efforts for other underrepresented language groups, demonstrating how each design decision (language addition, corpus balancing, preprocessing strategy) contributes quantifiable improvements on a defined evaluation framework. Fourth, the analysis of downstream impact on NanoGPT training demonstrates that tokenizer improvements translate directly to measurable gains in training loss, vocabulary utilization, and inference efficiency, establishing the practical significance of tokenization research for deployed educational AI systems.

Future work will pursue four directions: integrating sandhi-aware preprocessing for Sanskrit using the SanskritShala pipeline~\citep{sandhan2023sanskritshala}, extending vocabulary coverage to all 22 scheduled Indian languages through a tiered vocabulary strategy~\citep{gren2026efficient}, conducting comprehensive downstream task evaluations across question answering, translation, and summarization benchmarks for all seven currently supported languages, and investigating dynamic vocabulary adaptation techniques that could eliminate the static allocation trade-off by adjusting vocabulary composition at inference time based on detected input language.

The BHARATI tokenizer models, training scripts, and evaluation benchmarks are released under open licenses to support further research in multilingual Indian NLP and educational AI applications that serve diverse linguistic communities.

\section*{Funding}
This work was supported by the Institute of Eminence Funds (IM00002G\_RB\_SG), Planning Unit, IIT Delhi; Centre for SeNSE, IIT Delhi new faculty support; and Google Cloud research credits.

\section*{Acknowledgements}
The authors acknowledge computational resources of the Intelligent Robotics and Rebooting Computing Chip Design (INTRINSIC) Laboratory, Centre for SeNSE, Indian Institute of Technology Delhi, IM00002G\_RB\_SG IoE Fund Grant (NFSG), Indian Institute of Technology Delhi.

\section*{Conflict of Interest}
The authors declare no conflicts of interest.

\section*{Data Availability}
The BHARATI tokenizer models (v1, v2, v3), training corpora metadata, preprocessing scripts, fertility evaluation benchmarks, and all configuration files are available at \url{https://huggingface.co/RSL-INTRINSICLab-IIT}. The training corpus itself comprises publicly available data from the Sangraha collection, Project Gutenberg, and digitized classical texts.

\bibliography{refs_v2}

\begin{thebibliography}{42}
\providecommand{\natexlab}[1]{#1}
\providecommand{\url}[1]{\texttt{#1}}
\expandafter\ifx\csname urlstyle\endcsname\relax
  \providecommand{\doi}[1]{doi: #1}\else
  \providecommand{\doi}{doi: \begingroup \urlstyle{rm}\Url}\fi

\bibitem[Sennrich et~al.(2016)Sennrich, Haddow, and Birch]{sennrich2016neural}
Rico Sennrich, Barry Haddow, and Alexandra Birch.
\newblock Neural machine translation of rare words with subword units.
\newblock In \emph{Proceedings of the 54th Annual Meeting of the Association
  for Computational Linguistics}, pages 1715--1725. Association for
  Computational Linguistics, 2016.
\newblock \doi{10.18653/v1/P16-1162}.

\bibitem[Devlin et~al.(2019)Devlin, Chang, Lee, and Toutanova]{devlin2019bert}
Jacob Devlin, Ming-Wei Chang, Kenton Lee, and Kristina Toutanova.
\newblock {BERT}: Pre-training of deep bidirectional transformers for language
  understanding.
\newblock In \emph{Proceedings of NAACL-HLT 2019}, pages 4171--4186.
  Association for Computational Linguistics, 2019.

\bibitem[Kudo and Richardson(2018)]{kudo2018sentencepiece}
Taku Kudo and John Richardson.
\newblock {S}entence{P}iece: A simple and language independent subword
  tokenizer and detokenizer for neural text processing.
\newblock In \emph{Proceedings of the 2018 Conference on Empirical Methods in
  Natural Language Processing: System Demonstrations}, pages 66--71.
  Association for Computational Linguistics, 2018.
\newblock \doi{10.18653/v1/D18-2012}.

\bibitem[Petrov et~al.(2024)Petrov, La~Malfa, Torr, and
  Bibi]{petrov2024language}
Aleksandar Petrov, Emanuele La~Malfa, Philip~H.S. Torr, and Adel Bibi.
\newblock Language model tokenizers introduce unfairness between languages.
\newblock \emph{arXiv preprint arXiv:2305.15425}, 2024.

\bibitem[Goyal and Huet(2012)]{goyal2012distributed}
Pawan Goyal and G{\'e}rard Huet.
\newblock A distributed platform for {S}anskrit processing.
\newblock In \emph{Proceedings of COLING 2012}, pages 1011--1028. Association
  for Computational Linguistics, 2012.

\bibitem[Wu et~al.(2016)Wu, Schuster, Chen, Le, Norouzi, Macherey, Krikun, Cao,
  Gao, Macherey, et~al.]{wu2016google}
Yonghui Wu, Mike Schuster, Zhifeng Chen, Quoc~V. Le, Mohammad Norouzi, Wolfgang
  Macherey, Maxim Krikun, Yuan Cao, Qin Gao, Klaus Macherey, et~al.
\newblock Google's neural machine translation system: Bridging the gap between
  human and machine translation.
\newblock \emph{arXiv preprint arXiv:1609.08144}, 2016.

\bibitem[Radford et~al.(2019)Radford, Wu, Child, Luan, Amodei, and
  Sutskever]{radford2019language}
Alec Radford, Jeffrey Wu, Rewon Child, David Luan, Dario Amodei, and Ilya
  Sutskever.
\newblock Language models are unsupervised multitask learners.
\newblock \emph{OpenAI blog}, 2019.

\bibitem[G.(2025)]{abudouwailig2025research}
Abudouwaili G.
\newblock Research on morphological knowledge-guided low-resource agglutinative
  languages-chinese translation.
\newblock \emph{Complex and Intelligent Systems}, 11, 2025.
\newblock \doi{10.1007/s40747-025-01780-5}.

\bibitem[A.(2026{\natexlab{a}})]{phadtea2026analysing}
Phadte A.
\newblock Analysing unified embedding with morphological insight for
  multilingual text representation.
\newblock \emph{Engineered Science}, 40, 2026{\natexlab{a}}.
\newblock \doi{10.30919/es2102}.

\bibitem[Conneau et~al.(2020)Conneau, Khandelwal, Goyal, Chaudhary, Wenzek,
  Guzm{\'a}n, Grave, Ott, Zettlemoyer, and Stoyanov]{conneau2020unsupervised}
Alexis Conneau, Kartikay Khandelwal, Naman Goyal, Vishrav Chaudhary, Guillaume
  Wenzek, Francisco Guzm{\'a}n, Edouard Grave, Myle Ott, Luke Zettlemoyer, and
  Veselin Stoyanov.
\newblock Unsupervised cross-lingual representation learning at scale.
\newblock In \emph{Proceedings of the 58th Annual Meeting of the Association
  for Computational Linguistics}, pages 8440--8451. Association for
  Computational Linguistics, 2020.

\bibitem[Kunchukuttan et~al.(2020)Kunchukuttan, Kakwani, Golla, N.C.,
  Bhattacharyya, Khapra, and Kumar]{kunchukuttan2020ai4bharat}
Anoop Kunchukuttan, Divyanshu Kakwani, Satish Golla, Gokul N.C., Avik
  Bhattacharyya, Mitesh~M. Khapra, and Pratyush Kumar.
\newblock Ai4bharat-indicnlp corpus: Monolingual corpora and word embeddings
  for indic languages.
\newblock In \emph{Proceedings of the Twelfth Language Resources and Evaluation
  Conference}, pages 4948--4961. European Language Resources Association, 2020.

\bibitem[Kakwani et~al.(2020)Kakwani, Kunchukuttan, Golla, N.C., Bhattacharyya,
  Khapra, and Kumar]{kakwani2020indicnlpsuite}
Divyanshu Kakwani, Anoop Kunchukuttan, Satish Golla, Gokul N.C., Avik
  Bhattacharyya, Mitesh~M. Khapra, and Pratyush Kumar.
\newblock Indicnlpsuite: Monolingual corpora, evaluation benchmarks and
  pre-trained multilingual language models for indian languages.
\newblock In \emph{Findings of EMNLP 2020}, pages 4948--4961. Association for
  Computational Linguistics, 2020.

\bibitem[Kanjirangat(2025)]{kanjirangat2025tokenization}
Vani Kanjirangat.
\newblock Tokenization and representation biases in multilingual models on
  dialectal nlp tasks.
\newblock \emph{Proceedings of the 2025 Conference on Empirical Methods in
  Natural Language Processing}, pages 24003--24021, 2025.
\newblock \doi{10.18653/v1/2025.emnlp-main.1224}.

\bibitem[Nyalang(2025)]{nyalang2025krenne}
Badal Nyalang.
\newblock Kren-ne: A multilingual tokenization framework for northeast indian
  languages.
\newblock \emph{TechRxiv Preprint}, 2025.
\newblock \doi{10.36227/techrxiv.176184636.69664258/v1}.

\bibitem[Gu et~al.(2021)Gu, Tinn, Cheng, Lucas, Usuyama, Liu, Naumann, Gao, and
  Poon]{gu2021domain}
Yu~Gu, Robert Tinn, Hao Cheng, Michael Lucas, Naoto Usuyama, Xiaodong Liu,
  Tristan Naumann, Jianfeng Gao, and Hoifung Poon.
\newblock Domain-specific language model pretraining for biomedical natural
  language processing.
\newblock \emph{ACM Transactions on Computing for Healthcare}, 3\penalty0
  (1):\penalty0 1--23, 2021.
\newblock \doi{10.1145/3458754}.

\bibitem[Chalkidis et~al.(2020)Chalkidis, Fergadiotis, Malakasiotis, Aletras,
  and Androutsopoulos]{chalkidis2020legal}
Ilias Chalkidis, Manos Fergadiotis, Prodromos Malakasiotis, Nikolaos Aletras,
  and Ion Androutsopoulos.
\newblock {LEGAL-BERT}: The muppets straight out of law school.
\newblock In \emph{Findings of EMNLP 2020}, pages 2898--2904. Association for
  Computational Linguistics, 2020.

\bibitem[Lewis et~al.(2020)Lewis, Ghazvininejad, Ghosh, Celikyilmaz, and
  Zettlemoyer]{lewis2020pretrained}
Mike Lewis, Marjan Ghazvininejad, Gargi Ghosh, Asli Celikyilmaz, and Luke
  Zettlemoyer.
\newblock Pre-training via paraphrasing.
\newblock In \emph{Advances in Neural Information Processing Systems},
  volume~33, pages 18470--18481, 2020.

\bibitem[Li and Xing(2019)]{li2019word}
Chao Li and Ji~Xing.
\newblock Word segmentation for classical {C}hinese: New standards and a study
  on using pre-training.
\newblock In \emph{Proceedings of the Sixth Workshop on NLP for Similar
  Languages, Varieties, and Dialects}, pages 174--182. Association for
  Computational Linguistics, 2019.

\bibitem[Limisiewicz(2023)]{limisiewicz2023tokenization}
Tomasz Limisiewicz.
\newblock Tokenization impacts multilingual language modeling: Assessing
  vocabulary allocation and overlap across languages.
\newblock \emph{Findings of the Association for Computational Linguistics: ACL
  2023}, pages 5661--5681, 2023.
\newblock \doi{10.18653/v1/2023.findings-acl.350}.

\bibitem[Samuel(2023)]{samuel2023tokenization}
David Samuel.
\newblock Tokenization with factorized subword encoding.
\newblock \emph{Findings of the Association for Computational Linguistics: ACL
  2023}, pages 14143--14161, 2023.
\newblock \doi{10.18653/v1/2023.findings-acl.890}.

\bibitem[Chizhov(2024)]{chizhov2024bpe}
Pavel Chizhov.
\newblock Bpe gets picky: Efficient vocabulary refinement during tokenizer
  training.
\newblock \emph{Proceedings of the 2024 Conference on Empirical Methods in
  Natural Language Processing}, pages 16587--16604, 2024.
\newblock \doi{10.18653/v1/2024.emnlp-main.925}.

\bibitem[Salve(2025)]{salve2025tokenization}
Hyacinth~Nathalie Salve.
\newblock Tokenization efficiency in code-switched text: Comparing
  sentencepiece and byte-pair encoding on taglish.
\newblock \emph{TechRxiv Preprint}, 2025.
\newblock \doi{10.36227/techrxiv.175756338.86755792/v1}.

\bibitem[P.(2025)]{pattnayakp2025tokenization}
Pattnayak P.
\newblock Tokenization matters: Improving zero-shot ner for indic languages.
\newblock \emph{IEEE International Conference on Electro Information
  Technology}, pages 456--462, 2025.
\newblock \doi{10.1109/eIT64391.2025.11103625}.

\bibitem[Lian(2025)]{lian2025scaffoldbpe}
Haoran Lian.
\newblock Scaffold-bpe: Enhancing byte pair encoding for large language models
  with simple and effective scaffold token removal.
\newblock \emph{Proceedings of the AAAI Conference on Artificial Intelligence},
  39:\penalty0 24539--24548, 2025.
\newblock \doi{10.1609/aaai.v39i23.34633}.

\bibitem[Sandhan(2023)]{sandhan2023sanskritshala}
Jivnesh Sandhan.
\newblock Sanskritshala: A neural sanskrit nlp toolkit with web-based interface
  for pedagogical and annotation purposes.
\newblock \emph{Proceedings of the 61st Annual Meeting of the Association for
  Computational Linguistics (Volume 3: System Demonstrations)}, pages 103--112,
  2023.
\newblock \doi{10.18653/v1/2023.acl-demo.10}.

\bibitem[U.(2026)]{tukeyevu2026morphologyaware}
Tukeyev U.
\newblock Morphology-aware segmentation and tokenization for turkic languages:
  A cse-guided framework (the kazakh case).
\newblock \emph{Information Switzerland}, 17, 2026.
\newblock \doi{10.3390/info17020128}.

\bibitem[Mohamed(2025)]{mohamed2025multilingual}
Azhar Mohamed.
\newblock Multilingual tokenization efficiency in large language models: A
  study on indian languages.
\newblock \emph{SSRN Electronic Journal}, 2025.
\newblock \doi{10.2139/ssrn.5201887}.

\bibitem[S.(2025{\natexlab{a}})]{yadavs2025a}
Yadav S.
\newblock A preliminary exploration of phrase-based smt and multi-bpe
  segmentations through concatenated tokenised corpora for low-resource indian
  languages.
\newblock \emph{Conference on Machine Translation Proceedings}, pages
  1253--1258, 2025{\natexlab{a}}.
\newblock \doi{10.18653/v1/2025.wmt-1.103}.

\bibitem[A.(2025)]{madhavaraja2025subword}
Madhavaraj A.
\newblock Subword dictionary learning and segmentation for expanding the
  vocabulary of automatic speech recognition in tamil and kannada.
\newblock \emph{ACM Transactions on Asian and Low Resource Language Information
  Processing}, 24, 2025.
\newblock \doi{10.1145/3705312}.

\bibitem[Nehrdich(2024)]{nehrdich2024one}
Sebastian Nehrdich.
\newblock One model is all you need: Byt5-sanskrit, a unified model for
  sanskrit nlp tasks.
\newblock \emph{Findings of the Association for Computational Linguistics:
  EMNLP 2024}, pages 13742--13751, 2024.
\newblock \doi{10.18653/v1/2024.findings-emnlp.805}.

\bibitem[Gren(2026)]{gren2026efficient}
Gustaf Gren.
\newblock Efficient low-resource language models using tokenizer transfer.
\newblock \emph{Proceedings of the 19th Conference of the European Chapter of
  the Association for Computational Linguistics (Volume 4: Student Research
  Workshop)}, pages 639--648, 2026.
\newblock \doi{10.18653/v1/2026.eacl-srw.49}.

\bibitem[Sen(2025)]{sen2025architectural}
Arnab Sen.
\newblock Architectural evaluation of subword tokenization and compact language
  models (clms) for resource-constrained nlp deployment.
\newblock \emph{International Journal of Innovative Science and Research
  Technology}, pages 602--609, 2025.
\newblock \doi{10.38124/ijisrt/25nov578}.

\bibitem[Andryushchenko(2025)]{andryushchenko2025evaluating}
Georgy Andryushchenko.
\newblock Evaluating tokenizer adaptation methods for large language models on
  low-resource programming languages.
\newblock \emph{Proceedings of the 63rd Annual Meeting of the Association for
  Computational Linguistics (Volume 4: Student Research Workshop)}, pages
  823--833, 2025.
\newblock \doi{10.18653/v1/2025.acl-srw.57}.

\bibitem[Liu et~al.(2020)Liu, Gu, Goyal, Li, Edunov, Ghazvininejad, Lewis, and
  Zettlemoyer]{liu2020multilingual}
Yinhan Liu, Jiatao Gu, Naman Goyal, Xian Li, Sergey Edunov, Marjan
  Ghazvininejad, Mike Lewis, and Luke Zettlemoyer.
\newblock Multilingual denoising pre-training for neural machine translation.
\newblock In \emph{Transactions of ACL}, volume~8, pages 726--742, 2020.

\bibitem[Držík(2025)]{drk2025dictionarybased}
Dávid Držík.
\newblock Dictionary-based byte-pair encoding tokenizer for morphologically
  rich languages.
\newblock \emph{2025 International Conference on Electrical and Computer
  Engineering Researches (ICECER)}, pages 1--7, 2025.
\newblock \doi{10.1109/icecer65523.2025.11401129}.

\bibitem[Osterrieder(2025)]{osterrieder2025nlp}
Joerg Osterrieder.
\newblock Nlp - tokenization and subword models.
\newblock \emph{IDA-CL Working Papers}, 2025.
\newblock \doi{10.24818/ida-cl/2025.77}.

\bibitem[S.(2025{\natexlab{b}})]{narzarys2025bpe}
Narzary S.
\newblock Bpe and morphologically segmented phrase based statistical machine
  translation system for indian languages to resource constrained language
  bodo.
\newblock \emph{Multimedia Tools and Applications}, 84:\penalty0 29715--29732,
  2025{\natexlab{b}}.
\newblock \doi{10.1007/s11042-024-20277-w}.

\bibitem[García-Sierra et~al.(2025)García-Sierra,
  Fernández-Pampillón~Cesteros, and Ortega-Martín]{garcasierra2025a}
Óscar García-Sierra, Ana Fernández-Pampillón~Cesteros, and Miguel
  Ortega-Martín.
\newblock A morphological tokenizer for generating vocabularies for large
  language models in spanish.
\newblock \emph{Procesamiento del Lenguaje Natural}, 75:\penalty0 29--40, 2025.
\newblock URL
  \url{http://journal.sepln.org/sepln/ojs/ojs/index.php/pln/article/view/6736}.

\bibitem[Yadav(2026)]{yadav2026maibert}
Sumit Yadav.
\newblock Maibert: A pre-training corpus and language model for low-resourced
  maithili language.
\newblock \emph{Proceedings of the Second Workshop on Language Models for
  Low-Resource Languages (LoResLM 2026)}, pages 444--452, 2026.
\newblock \doi{10.18653/v1/2026.loreslm-1.38}.

\bibitem[A.(2026{\natexlab{b}})]{reyesveraa2026spadebert}
Reyes-Vera A.
\newblock Spade-bert: Multilingual bert-based model with trigram-sensitive
  tokenization, tuned for depression detection in spanish texts.
\newblock \emph{AI Switzerland}, 7, 2026{\natexlab{b}}.
\newblock \doi{10.3390/ai7020048}.

\bibitem[Manukonda(2024)]{manukonda2024enhancing}
Durga~Prasad Manukonda.
\newblock Enhancing multilingual natural language processing with custom
  subword tokenization: Subword2vec and bilstm integration for lightweight and
  streamlined approaches.
\newblock \emph{2024 6th International Conference on Natural Language
  Processing (ICNLP)}, pages 366--371, 2024.
\newblock \doi{10.1109/icnlp60986.2024.10692344}.

\bibitem[Erkaya(2023)]{erkaya2023analysis}
Erencan Erkaya.
\newblock Analysis of subword tokenization approaches for turkish language.
\newblock \emph{2023 31st Signal Processing and Communications Applications
  Conference (SIU)}, pages 1--4, 2023.
\newblock \doi{10.1109/siu59756.2023.10223973}.

\end{thebibliography}

\end{document}